# Transient Diversity in Multi-Agent Systems

David Lybäck

Royal Institute of Technology, Sweden

Diversity is an important aspect of highly efficient multi-agent teams. We introduce the main factors that drive a multi-agent system in either direction along the diversity scale. A metric for diversity is described, and we speculate on the concept of transient diversity. Finally, an experiment on social entropy using a RoboCup simulated soccer team is presented.

This research area is quite new, but we are assured that improved diversity management for multi-agent systems can make immediate contributions to existing agent-based applications as well as to long-range development in future multi-robot technologies.

Categories and Subject Descriptors: I.2.11 [Artificial Intelligence]: Distributed Artificial Intelligence - *Intelligent Agents*.

*Contents*

# 1 INTRODUCTION

*'Let every eye negotiate for itself,
And trust no agent.'* [1]

What is the magic factor for achieving effective teamwork? Is it a matter of dynamic coordination between the team members, or is it to decide on some sort of clever delegation of tasks and skills?

We believe that a bounded degree of agent specialization, resulting in a bounded heterogeneity, is important when trying to achieve effective task completion, but that there must be team coherency factors in effect as well. We will therefore suggest a structure that integrates the factors that influence the overall diversity of a multi-agent system. The proposed factors drive the system in either of two directions on the diversity scale: towards heterogeneity or towards homogeneity, and we will therefore call them diversity drivers. In our model, the diversity drivers, taken together, constitute a diversifying force on the system, varying over time, and towards either end of the diversity scale.

We will discuss suggested metrics for diversity. Without a quantitative metric of the extent of diversity in a system, we will not be able to correlate diversity with performance.

Finally, we will introduce a concept of transient diversity in artificial systems, and thereby explain how a diversity mechanism meant to achieve adaptive and efficient teamwork may also include a risk of counter-effective behavioral oscillations.

## 1.1 BACKGROUND

### 1.1.1 AGENTS

As a further development of program objects, agents[2] react according to the actual situation, taking into account individual and group objectives. The notion of an agent is a very natural metaphor, and can be used as a key abstraction in many systems (Jennings and Woolridge, 1998).

The idea of agents will hopefully, via its reducing of complexity in systems modeling, contribute to the realization of advanced robotics and intuitive computer systems, and thus contribute to a more harmonious integration of information technology with human society (Watt, 1996).

### 1.1.2 TEAMWORK

Teamwork in multi-agent systems is an important issue, because we believe that good teamwork can be fundamental for efficient performance in many domains. Teamwork is an issue in e.g. the RoboCup challenge (Kitano et al, 1997). Researchers are trying to handle the complexity of the interactions of the agents by finding heuristics, and effective ways to model societies of agents. Some researchers are exploring social exchange using game theory (Cohen, Riolo, and Axelrod, 1998). We will consider concepts like traits and roles as ways to simplify and speed up the decision process in a complex real-time teamwork situation.

---

[1] William Shakespeare (1564-1616), English dramatist, poet. Claudio, in Much Ado About Nothing, act 2, sc. 1.
[2] For an overview of the agent concept, see e.g. (Engström and Kummeneje, 1997).

### 1.1.3 DECISIONS

In complex and dynamic domains, it is not possible to optimally plan in advance all necessary coordinated activities. The search space quickly becomes unmanageable, since the fast changing environment makes the search space extremely huge, and unexpected events in the environment makes the search space incomplete. Instead, all team members must continuously reflect upon the team's long-term expected utilities in the current situation. Thus, a long-term team can refine its *modus operandi* via self-reflection. A nice overview on teamwork in persistent teams is (Tambe and Zhang, 1998). Here, the authors point out the differences between a temporary coalition of agents, and the issues facing a persistent team. The key is to find an efficient operationalization of teamwork. If the model is simple, decision support can be received fast enough (Younes, 1998). It is worth noting that a full search of all possible future states, and their utilities, is practically impossible. Therefore, heuristics and an any-time decision method must be constructed. This is particularly important in a hostile or competitive situation, where deliberation must be fast.

### 1.1.4 TASK ADAPTION

The teams of interest to this thesis will act in complex domains and meet more or less unexpected challenges. Many properties of the team and of the individual agents can therefore not be set beforehand, but must be adaptive to the task at hand. We believe that task adaptive decisions by a team can directly affect one or more of the factors included in the diversity drivers under study, and will show this below.

### 1.1.5 WAIVER

Finally, it must be stressed that the systems under study are purely artificial ones. Whether or not principles and results in this field may be applicable to humans, human organization, or natural systems is of no primary concern at this point.

## 1.2 OBJECTIVES

The objectives of this project were to:
1. Investigate the concept of heterogeneity in multi-agent systems.
2. Investigate individual traits and roles as part of what heterogeneity stems from.
3. Participate in the software development of agent features and teamwork functionality within the department's RoboCup simulation league effort, the program system UBU.
4. Experiment on various degrees of heterogeneity in UBU and relate this to a performance metric.

## 1.3 METHODOLOGY

*A work is never completed except by some accident such as*
*weariness, satisfaction, the need to deliver, or death:*
*for, in relation to who or what is making it,*
*it can only be one stage in a series of inner transformations.[3]*

After an initial literature survey, the project work focused on objective 3 (software development) for a two months. A three-week study journey to Japan was made, including participation in RoboCup Japan Open, and meetings with researchers in the field. Thereafter investigation of traits, roles, and diversity began, removing trait studies as a main goal, and instead broadening objectives 1 and 2 to a general understanding and classification of diversity in multi-agent systems.

Participation in the RoboCup World Cup 99 and the Sixteenth International Joint Conferences on Artificial Intelligence was a good milestone towards the end of the project.

Finally, a small experiment was performed to measure the effects on player positioning entropy vs performance in the UBU soccer team. A late planned extra objective of design and implementation for experiments on transient diversity was canceled due to lack of time.

It must be stressed that the small experiment performed is *not* an empirical support to the theories presented in this thesis, especially not to the speculations presented in chapter 5. We hope that this thesis will nevertheless stimulate additional research in the field.

## 1.4 RELATED WORK

### 1.4.1 SOCIAL ORGANIZATION IN INSECT-LIKE SOCIETIES

According to the theory of evolution, the complexity in the environment drives the adaptation of the species. Research in societies of numerous simple insect-like creatures have been made in search for an emergent "swarm intelligence", but we are mainly interested in systems with real mobile robots that have been used for experiments on cooperation methodologies. Maja J Mataric at the Brandeis University built a multi-robot foraging system to explore the basics of social learning (Mataric, 1997). In this architecture, basic behaviors serve as a substrate for synthesizing into higher-level group interactions. In the experiment, greedy agents learn a social rule that benefits the group as a whole. The social rule of turn taking, by yielding and sharing information, actually evolved in the system.

Interestingly, Mataric uses three forms of so-called social reinforcements, which also fit smoothly into the homogeneity drivers presented in this thesis. The first one is an *individual judgment of progress* relative to the current goal. (This corresponds to the local reinforcement driver presented on page 14). The second reinforcement is *observation of the behavior of conspecifics*. (This corresponds to the imitation factor presented on page 13). Finally, the third type of social reinforcement in Mataric's multi-robot system is *reinforcement received by conspecifics* (This corresponds to the upholding of norms and agreements presented on page 10 onwards, and the vicarious feedback factor presented on page 13).

---

[3] Paul Valéry (1871–1945), French poet, essayist. "Recollection" (published in *Collected Works, vol. 1*, 1972). From The Columbia Dictionary of Quotations.

1.4.2 BEHAVIORAL DIVERSITY IN LEARNING ROBOT TEAMS

In his Ph.D. thesis, Tucker Balch focused on behavioral diversity and learning in multi-robot societies, primarily in the domains of multi-robot foraging and robotic soccer. He has also worked with formation control, which allows cooperative motion of robots relative to each other (Balch, 1998). His presentation of diversity metrics has been of profound importance for this project, and is heavily referred to on pages 15-20.

## 1.5 STRUCTURE OF THESIS

Chapter 2 proposes four main heterogeneity drivers. Chapter 3 proposes four corresponding homogeneity drivers. Chapter 4 describes diversity, and chapter 5 introduces the concept of transient diversity. Chapter 6 overviews the experimental platform, lists the results from the experiment, and discusses the results. Related work is also presented. Chapter 7 concludes the thesis, and points out future work.

# 2  HETEROGENEITY DRIVERS

*We cannot feel strongly toward the totally unlike because it is unimaginable, unrealizable;*
*nor yet toward the wholly like because it is stale—identity must always be dull company.*
*The power of other natures over us lies in a stimulating difference which causes excitement and opens communication,*
*in ideas similar to our own but not identical, in states of mind attainable but not actual.*[4]

At least four main factors, or drivers, push the multi-agent system towards heterogeneity. Of these, only the first driver includes the traditional engineering aspects. Two of the remaining drivers are a fusion of several fields into the multi-agent programming domain. The last driver is connected to recent findings from research using artificial neural networks for multi-agent systems.

## 2.1  TECHNICAL DIFFERENCES

The hardware and software may vary between the individual agents in the team. Some variations are designed features, and some are drawbacks because of practical necessities, e.g. available resources. We will briefly mention a few typical hardware and software differences that arise between individual agents. The list depends on the actual system under study, and can probably be made longer.

### 2.1.1 HARDWARE

The technical hardware differences are mostly about the physical construction, the computer hardware, and the network facilities.

*Physical Construction (Robotics)*

The physical construction, i.e. mechanical and electromechanical devices, traction, arms, gripping devices, etc., varies between robot models. There is also variation between individuals of the same model due to minor manufacturing differences, maintenance done, and configuration changes made on the individual since manufactured.

At first glance, this factor seems very static when evaluating a multi-agent system, but it is not necessarily so. For certain robot models, a great deal of design effort will be put on versatility through multi-purpose connectors, adapters, and 'plug-and-play' physical extensions (Fujita, 1999). Some robots will then allow the exchange of functionality modules during operation, either via self-performed re-configuration, or via docking to a re-equip bay of some sort. Hence, the physical construction factor can be dynamic during robot operation.

*Computer Hardware*

The available computing resources have a profound impact on the performance of the agent. Many researchers work hard on minimizing the need of computer resources in critical run-time situations, by using so-called off-line time to perform a lot of preparative computing, by optimizing algorithms, and by setting priority-based execution for concurrent threads.

---

[4] Charles Horton Cooley (1864–1929), U.S. sociologist. *Human Nature and the Social Order*, ch. 4 (1902). From The Columbia Dictionary of Quotations.

*Network*

The availability of network facilities, and the actual performance of the network (the quality of service), is a substantial factor for many agents when it comes to run-time technical differences. The three most important quality of service items, thus influencing the network heterogeneity factor, are throughput, delay, and reliability (Reichert and Maguire, 1994).

Imagine if one of the agents can connect easily to an important server, and respond just before all other agents, or if one agent have much less problems with lost messages, then there is reason to believe that this agent might have an advantage. Note that this is the case even if the system is coordinated at discrete time steps with full information before decisions guaranteed. A loss of processing time is a drawback, even if all information eventually arrives to the agent, since processing time is a critical resource in real-time domains.

In a few cases, all agents in the system are assigned the same resources, via the same or similar connections. In those cases, if network resources have a high availability (almost no congestion), allocations are swift and smooth, and all arbitration of resources between the agents is fair, then the network differences need not be a substantial factor in the analysis of heterogeneity drivers.

### 2.1.2 SOFTWARE

The software can vary due to the actual program system in use, the current build configuration of the components, and the technical start parameters given to the agent program. Sometimes the configuration is a desired and expected feature, sometimes differences arise more or less unexpectedly after platform negotiations, operating system resource allocation control, etc.

## 2.2 ROLE ASSIGNMENT

Individuals in many multi-agent systems specialize. Sometimes this occurs simply because the domain includes rules concerning roles. Such rules state how role assignment promote or limit various options for the agent.

Consider for instance the goalkeeper in a soccer team. It would be unfruitful for a soccer team if no member used the available option to catch the ball using hands. So, all teams have a goalkeeper doing exactly this, to effectively intercept in 'dangerous' situations. It would even be nice for a team to have two goalkeepers. However, in soccer, there is also the well-known rule that only one individual is allowed to perform the catch operation, viz. the appointed goalkeeper. If this rule is not adhered to, a penalty kick is given to the other team, which implies a severe risk of losing a score, and maybe thereby in the end losing the game. So, the soccer domain promotes a specialization into at least the two roles of goal keeper and field player. An overview of roles in robotic soccer is (Åberg, 1998). Decision support for the RoboCup challenge is reviewed in (Åhman, 1998).

### 2.2.1 MANDATORY ROLE ASSIGNMENT

In some contexts, there might be a law in effect that one individual in a group must bear a special responsibility, e.g. captain of a ship, pilot-in-command of an airplane, or president of a corporation. Then it would only be futile refraining from designating team members into required roles. Imagine having the officers of a ship deciding not to use a captain, or the employees of a corporation agreeing to stop having a president. No matter the arguments why the team makes such a decision, the outcome is more ridiculous than effective, because this kind of role assignment is demanded by the context in which the multi-agent system operates.

The formalization of the division of tasks into roles is often made well in advance, for safety reasons. Airline pilots should be very sure of the procedures before even starting the engines. Crisply agreed roles are especially useful in emergencies or in stressful circumstances, where there is no room for misunderstanding.

Sometimes mandatory role assignments, and stiff behavior definitions are overused, resulting in dysfunctions and organizational pathologies (see e.g. Mintzberg, 1983). Of questionable validity is of course formalized behavior based only on some general desire for order, like tennis players wearing white shirts. Whereas these rule-bounded assignments due to external demands are of great interest in law and organization theory, they are only peripheral to agent programming.

2.2.2 LEARNED ROLES FOR EFFECTIVE TEAMWORK

The other kind of role assignment is the self-organizing, or learned role assignments, used for effective teamwork. This is a central issue of many experiments in the multi-agent systems field. In comparison to the mandatory roles mentioned above, the learned roles generally have a more dynamic and fluid structure, especially initially, before relatively stable roles are assumed. The mandatory roles made by context rules, or laws, tend to be of a rigid nature. The learned roles are more of a free-will specialization of team members, where they engage in specialization to increase effectiveness and overall team performance. Think about how firemen act when reaching the scene of a fire. They don't start a lengthy discussion about who will attach the hoses and who will turn the valves for the hydrant. There are no arguments about who will go up the ladder depending on the day of the week or what-have-you. It is clear to any observer that the learned roles for the standard situation are improving the efficiency of the team immensely.

The border between suggested heterogeneity drivers is neither sharp, nor rigid. Learned roles can be transformed into mandatory roles, and back. The tradeoff between individual and global efficiency achieved via roles can be one of the engineered settings for the artificial agents, and thus the role factor also has a connection to the technical difference factor, etc.

## 2.3 INDIVIDUAL TRAITS AND SKILLS

2.3.1 LEARNING SKILLS

*Off-line learning* is when training examples are processed until the learned behaviors are good enough. A drawback is that learning takes a long time, inasmuch as a large number of examples must be processed, so that a specific situation and task combination reoccur often enough to enable learning opportunities. The off-line learning must also take into account that the capabilities of other agents can be quite different in the real operating conditions compared to the training examples.

In real-time conditions, an agent can then tailor specific skills according to the role and situations faced: the so-called *on-line learning*. One difficulty is that applying e.g. reinforcement learning for the on-line learning does not adapt rapidly enough. A remedy to this is to apply limited social learning (see page 8). Thus, the exact functionality of the agent is not fixed from the start, but rather continuously refined as a result of intensive interaction with its environment, as noted by (Maes, 1990).

2.3.2 INDIVIDUAL TRAITS

A trait is a predisposition to act in the same way in a wide range of situations (Hjelle and Ziegler,

1992). Traits are statements of:

1) Probability of certain behavior in response to particular situations.
2) Rates of change in above-mentioned probability.

So, in more scientific terms, a trait is a differential response bias. It is not a simple coupling between just one specific stimulus and one specific response. Instead, a trait is more of a broad consistency in behavior, cross-situational, and often quite stable.

Psychologists disagree not only about the existence and relevance of traits in humans, but also about the number and name of the basic traits. This discussion is omitted from this thesis, but an excellent introduction to trait theory is (Matthews and Deary, 1998). In humans, there is also some intervention between situation, traits, and response: emotions and cognitive states. For an overview of personality in artificial characters, see (Rousseau and Hayes-Roth, 1996).

The dispositional perspective in personality theory, with researchers such as Gordon Allport, Raymond Cattell, and Hans Eysenck, where the notion of traits is central, is often treated separately from the social cognitive perspective, represented by Albert Bandura and Julian Rotter, as well as from the cognitive perspective of George Kelly. However, a trait can also be seen as an example of a construct used as a cognitive template, so referring to trait theory must not necessarily be a stance in the human psychology debate.

### 2.3.3 LOCALIZED SOCIAL LEARNING

At first, one might like an improved skill to spread to the other team members as well, but in realistic environments, there are limits to the amount of social learning activity possible to concurrently uphold. Some transfer is possible, but for communication limitations and other resource constraints, the skill transfer is usually low or non-existent. This means that the agents will refine their skills in different ways, leading to specialization (Tambe et al. 1999). Furthermore, results in various game simulations indicate that localized social learning leads to more cooperation than widespread learning strategies. A recent speculation is that by establishing a "shadow of the adaptive future", viz. the preservation of agent context over time, localized social learning "offers some additional opportunities to catch misleading lessons from noisy experiences, before they diffuse widely through the population" (Cohen, Riolo and Axelrod, 1998).

## 2.4 GLOBAL EVALUATION OF PERFORMANCE

Global reinforcement is when the same performance feedback signal is delivered to all agents in the team. This reinforcement method correlates correctly with the overall team performance, but it may in many situations confuse agents that they behaved correctly or wrongly, when they were not actively involved in the activity leading to the reinforcement. The misattribution inferences are a big problem in learning through global reinforcement (Balch, 1998).

A specific problem is that in many tasks a performance-based reward is delayed. This makes it difficult for the agent to assign credit or blame to actions. A suggested solution to this problem is to provide a heuristic reward that provides more immediate feedback.

We believe that, in general, a global reinforcement scheme will *not* make initially heterogeneous agents acting in a complex, dynamic domain, converge into a similar behavior for a long time, if at all. Therefore, global reinforcement is systematized in this thesis as a *weak* heterogeneity driver.

# 3   HOMOGENEITY DRIVERS

*Order is Heaven's first law; and this confessed,*
*Some are, and must be, greater than the rest,*
*More rich, more wise; but who infers from hence*
*That such are happier, shocks all common sense.*
*Condition, circumstance, is not the thing;*
*Bliss is the same in subject or in king.*[5]

At least four main factors, or drivers, push the multi-agent system towards homogeneity. Of these, only the first driver includes the traditional engineering aspects. Two of the remaining drivers are a fusion of several fields into the multi-agent programming domain. The last driver is connected to recent findings from research using artificial neural networks for multi-agent systems.

## 3.1 ENGINEERING CONSTRAINTS

### 3.1.1 DOMAIN CONSTRAINTS

Most multi-agent systems are developed for a specific domain, e.g. goods offering surveillance in e-commerce, multi-robot foraging mission, etc. This is the because of the typical engineering decision on level of modeling. Using an overly general agent design may complicate processing and even hinder smooth execution of the task at hand. The extra abstraction levels, protocols, and type checks necessary in an overly general approach may blur the vision for the programmers, and thus present a latent program error risk. Consequently, avoiding superfluous functionality can have a profound beneficial impact on performance and robustness of the system. To wisely abstract domain constraints into the system design early on in the design process is, in our opinion, of paramount importance to the success of the engineering part of a major software project.

### 3.1.2 DEVELOPMENT CONSTRAINTS

No complex technical system is complete in all aspects, or perfect. There are often a great number of theoretically possible situations that have not been fully analyzed, or even identified before the system is put into operation. Many compromises have to be made throughout the development phase, where the lack of time prohibits full development of adequate handling of all special cases, as well as puts aside various interesting opportunities envisioned by the design team.

*Project Resources*

The lack of resources when developing a program usually puts customization and specialization rather low on the priority list, at least compared to all the basic functionality needed. More often than not, the system has to be put in operation with many special cases not properly handled. When agents in a complex environment apply over-general methods, this equals insufficient specialization of roles and skills. It is theoretically possible to first implement a collection of overly specialized agents for a given set of tasks, then set the system in operation before the simplification process is completed. In practice, the agents are instead sub-classed from a simple, generic agent class, then improved upon and specialized, as far as the project constraints permit. Thus, development constraints generally drive the system towards the homogeneity end of the diversity scale.

---

[5] Alexander Pope (1688–1744), English satirical poet. *An Essay On Man,* Epistle 1. From The Columbia Dictionary of Quotations.

*Legacy*

In many engineering projects, there is an inheritance from previous work, in the form of components. Most development teams try to reuse well-functioning components into new projects if possible, to save time and trouble. Already written programs can be used as components in the developed system, and the code included in this fashion is sometimes called legacy software. If legacy software, instead of customized programs, is used to a great extent, this drives the system towards homogeneity.

We will now make a quick observation on how components can be installed in an agent system. An entity in an artificial system where there is inter-agent communication, is an agent only if it can communicate correctly according to a specified protocol, or Agent Communication Language (ACL). The ACL and its grammar can be very simple, but the communication would not be fruitful without it. How can existing programs (or rather sub-programs) be transferred into an agent program, adhering to the ACL in use?

There are three ways to do this: to implement a *transducer* that mediates and translates messages, to implement a *wrapper*, which is code injected on top of existing code to provide proper interface capabilities, or, finally, to *rewrite* the original program (Genesereth and Ketchpel, 1994).

## 3.2 NORMS AND AGREEMENTS

*'Never speak disrespectfully of Society, Algernon.*
*Only people who can't get into it do that'.* [6]

There are two main reasons to add a norm structure to the agent: safety aspects, and social functionality. To improve teamwork, various agreements can be made between the members of the team. There is a connection between norms and the role assignments mentioned above, since role compliance expectations could be formalized into a norm, etc.

### 3.2.1 NORMS

We will describe how norms can be used for safety, to facilitate social action, and to detect system abnormalities. We will present the maladaption criterion as a suggested base for norm construction.

*Norms For Safety*

All complex systems should include error detection and error handling. This is important for all software, to ensure a robust performance. In physical robots, this demand is even stronger. Without appropriate handling of exceptions, a faulty decision of a mobile robot in a dangerous situation could lead to catastrophic consequences, with the robot being totally destroyed. A robot acting in the physical world can also cause damage to property, nature or even harm humans or animals. Thus, the most important reason to add a small, but strict safety policy structure to the agent, is for safety reasons. For an introduction to the basics of harmful action avoidance, see (Weld and Etzioni, 1994).

---
[6] Oscar Wilde (1854–1900), Anglo-Irish playwright, author. Lady Bracknell, in The Importance of Being Earnest, act 4.

Boman (1999) suggested that the norms at hand are a very important safety policy structure. In Boman's suggested agent architecture, the norms either modify the decision bases, or filter out unwanted actions. There is a small architectural resemblance between Boman's suggestion for artificial agents and the superego imperatives filtering process acting in humans as hypothesized by Freud during the early 1920s (Hjelle and Ziegler, 1992). However, the Boman agent does not necessarily actively reason about the norms while making a rational decision.

*Norms to Facilitate Social Action*

In societies of agents, there is often unavoidable interference between the agents. The interference can have various manifestations, but comes in two types: resource competition and goal competition (Mataric, 1997). The competition for shared resources, whether it be space, provisions, or information, can lead to deadlocks or starvation. Goal competition is when agents due to lack of coordination undo the work of each other. Imagine for instance two robots wanting to move the same two items each at a time but to different location. When the first robot drop one of the items, the other one will take it and bring it back and *vice versa*. Unknowingly, they end up in a practical live-lock situation. In other cases, the goal competition has other manifestations, the essence being that the agents make life more difficult for each other for no reason. So, even if not actively cooperating, the agents from the same team should at least try to minimize their mutual interference.

It is quite obvious that without a model of social functions, collaboration would not only be shortsighted, purely reactive, but also fragile and unreliable. Instead, a social structure must emerge and stabilize at various efficient *modi operandi*, via some evolutionary process or learning methods (Castelfranchi, 1997). What is the key component of this social structure?

Norms may serve as the framework for a social structure. The learning of norms is a foundation for socially intelligent behavior, since the norms calibrate the autonomy level and prevent unwanted actions (Verhagen and Boman, 1999). Norms can originate from outside the system, but can also be formed within (Conte et al., 1998). Externally enforced norms are rules of the domain, often associated with a cost or penalty for violations. Internally formed group norms are policies within the group, best adhered to, but often without explicit punishment for deviation. Behavior sets are often formalized into roles to simplify teamwork, as mentioned on page 6, and there is often a connection between norms for social action on the one hand, and individual role descriptions with role compliance expectations on the other.

One way to define a group of agents is via the norms they are obeying. If an agent obey the norms of a group, it can be treated as a group member. Magnusson and Svonni (1999) have suggested a mathematical model of a multi-agent system with groups and norms.

*System Abnormalities*

There is a huge need, in contemporary and especially future multi-agent systems, of methods to single out abnormalities, not only for safety, but also for robust performance. In fault-tolerant software systems, a dead process should quickly be removed, and a new substitute immediately and automatically spawned. Erlang is a software system developed at the Ericsson research company Ellemtel in the late 1980s, supporting, among other things, so-called death notice subscriptions via double-direction process links, for sustained robust execution (Riboe, 1994). In Erlang, exit signals are sent from dying processes. In the multi-agent systems being studied in this thesis, the agent might be dying in another sense. A robot might be dying due to power failure or short-circuiting. Other kinds of abnormalities entered by an agent could be, e.g., engaging in a dangerous course of action, repeatedly performing strangely unproductive tasks, or simply severely interfering with a team member for no reason. Obviously, a malfunctioning physical robot cannot be replaced in one second by spawning a new one, but nevertheless, abnormalities must be detected, and actions taken accordingly.

### *Using Statistical data for Abnormality Detection*

A degenerate group norm is a norm simply made up of statistical inference data, where the median value is, without further ado, wrongly regarded to be the best possible value, and deviations treated as unwanted. (Note that the word 'abnormal' means 'away from the norm'). The problem with using statistical norms as a base for detecting abnormalities is obvious.

Think of a skillful goal-keeper using an original, but very successful positioning technique, and thus, based on statistical norms, being classified by the team, or the coach, as 'abnormal', thereby facing severe risk of being removed from the team. That would be a sub-optimal decision for the team. Hence, even though a multi-agent system can find use of statistical norms for abnormality detection, if error handling rules are thereby indelicately applied with brute force, we fear that they might do more harm than good for the overall performance of the system, as we saw in the previous example.

### *The Maladaption Criterion*

A functional group norm that prevents unwanted behavior should instead check for the *maladaption criterion* (Atkinson et al., 1990). According to this criterion, maladaptive behavior has adverse effects on the performance of the individual, or on the team as a whole.

Is there a connection between useful norms, and how members of the coalition regard the long-term utilities of various series of actions, thus formalizing the maladaption criterion? If that is the case, perhaps some useful norms actually originate from compiled agreed upon long-term team utility information. The useful norms would then be like crystallized group utility propositions, or accumulated experiences of previously successful adjustments. Acting based on useful norms[7], includes the advantage of reducing the complexity of decisions.

But the norms will only remain useful if uprising differences between norms and reality are reconciled. Thus, as long as the situation can be correctly assessed, and the prescribed pattern of actions is rational, the norms will be safeguarding controls, preventing dangerous mistakes, as well as upholding the amount of conformity needed for successful cooperation (Mannheim, 1950).

#### 3.2.2 COALITION AGREEMENTS

Often, the optimal course of action for an agent depends on that selected by another. If two or more agents pursue the same objective, there might be a delay or even more drastic outcomes, unless they coordinate. The existence of a coordination problem complicates the utility-based reasoning, seen from a general decision theoretic standpoint. Then, one might ask, what is the valuation for invoking a state where coordination is a potential problem? Well, the very essence of teamwork is to perform a series of coordinated actions to reach a shared goal.

Coordination and a series of more or less subtle interactions between the agents take place throughout a teamwork mission. All realistic teamwork conditions will provoke various failures to execute what was first intended: there may e.g. be misperceptions of what is really happening, and thereby faulty classification of the action of others.

---

[7] Some traditions and customs in the human society, are often transformed into social norms. Consider, e.g., the routines and expectations around important family events like weddings.

When the agents evolve differently because of learning, there may be some misjudgment of the experience of others, and how that may effect their decisions. Finally, because of all reasons mentioned, there may be errors in understanding the strategies or tactics chosen by the other team members.

As a remedy to this threat against efficient teamwork, the team may use norms for social action (as mentioned on page 11), but it can also dynamically adopt a multitude of coalition agreements. The coalition agreements typically have a looser compliance expectation than norms.

### *Coalition Agreements*

Long-term agreements often concerns strategy, role structure, and overall goal descriptions. As indicated by its name, these agreements usually have a long duration, but can be re-negotiated much quicker than norms. The long-term agreements might concern the whole team, or a semi-permanent sub-team, which can be called a *formation* or a *group*. In sports, long-term coalition agreements are often referred to as locker-room agreements (Stone et al., 1998). The long-term agreements can be transformed into norms via some norm formation process, which is out of scope of this thesis[8].

A short-term coalition agreement might be to perform a joint action, and thereafter dissolve. Such a coalition is sometimes referred to as a *unit* or a *task force* (the latter maybe sounding rather military).

### *Subjective Coalition*

In contrast to the overall objective coalition structures based on mathematical evaluation of norm adherence as mentioned above, a competing idea is the one of subjective coalition.

The main point in the subjective coalition concept is of course that the agents are autonomous and that they constantly build their own opinion of the state of the world. The social factors, e.g. 'who is cooperating with me on these issues', etc., are also matters of subjective judgment. The coalition can be formalized into a negotiated contract, but even after this formalization, the agents should continue to monitor the progress of the cooperation, and the mutual adherence or lack of adherence to the formalized coalition.

#### 3.2.3 VICARIOUS FEEDBACK

Socially intelligent behavior is connected to norm adaptation capacity (Boman and Verhagen, 1998). We believe that feedback from other team members is important to quickly achieve teamwork through coalitions or norm adherence. The vicarious feedback may complement individual evaluations of progress towards a goal. The vicarious feedback can also be the only way for the agent to effectively learn about ongoing norm building, and a major input channel for the cognitive process of norm integration.

## 3.3 IMITATION

*Those who do not want to imitate anything, produce nothing.[9]*
--
*Imitation, if it is not forgery, is a fine thing.*
*It stems from a generous impulse, and a realistic sense of what can and cannot be done.[10]*

---
[8] The need for transformation of an agreement into a norm might be connected to the cost of breaking the agreement. Some agreements should be kept loose, and not be transformed into a norm, since that might exaggerate conformity, and thus reduce team performance.
[9] Salvador Dali (1904-89), Spanish painter. "The Futuristic Dali" (1970). From The Columbia Dictionary of Quotations.

As mentioned on page 8, social learning can make the learning curve steeper, and facilitate the speedy learning of skills. On the other hand, social learning taken to the extreme is a matter of mere imitation. If the agents in the multi-agent system imitate each other (instead of independently exploring the domain and learning individual skills), there is reason to believe that they rapidly will eliminate all evolutionary variety of tactics and skills, and probably get stuck in inefficient, uncooperative behaviors. In rare cases, where the modes of cooperation are trivial, using imitation will nevertheless be a quick way to drive the multi-agent system towards sustained cooperative behavior, through the forceful homogeneity driving factor of imitation (Cohen, Riolo, and Axelrod, 1998).

### 3.4 LOCAL EVALUATION OF PERFORMANCE

The performance evaluation function (reward policy) in effect may make the agents choose similar egocentric policies, and this drives the system towards homogeneity. Note that this individuality through greediness is just another kind of homogeneity, apart from the flock effect induced by narrow norms, strict cooperation agreements, or pure imitation.

The information available to the agent is subjective in nature: all information is received via sensory input, even communication must be received through normal input layers. Furthermore, the agent is the acting entity.

This means that the rewards must be properly distributed over individual state/action pairs, or some similar structure, in order to make the agent learn which strategies that actually leads towards the goals, and which should be discarded. In many agent architectures, a local reward is received after an individual judgment of progress towards the current goal. It is well known that learning through local reinforcement works in a technical sense, because it provides a strong relation between the action and the reward/punishment, thus promoting a steeper learning curve than global reinforcement (Balch, 1998). But the team might not perform optimally; local reinforcement produces greedy agents, and thereby hinders the evolution of cooperative strategies. Incentives can easily become counterproductive (Chakravarthy and Lorange, 1991). So, although providing a steep learning curve, there is clearly a drawback with using local rewards.

On the other hand, some kind of incentives must be in effect, if a functioning system is to be created through learning. In a system of autonomous agents acting in noisy, dynamic conditions, where perception is limited, who then is to judge the activities undertaken by the agent, better than the agent itself? The individual agent is, after all, the only one that actually resides on all the information that is available in the specific situation, thereby presumably the one that can best judge the outcome of its local action. This dilemma has been studied by game theorists (Cohen et al., 1998), and especially by researchers training neural networks for cooperative action (Oota, 1999).

---

[10] James Fenton (b. 1949), British poet, critic. "Ars Poetica," no. 47, in Independent on Sunday (London, 16 Dec. 1990). From The Columbia Dictionary of Quotations.

# 4  DIVERSITY

*If there were only one religion in England there would be danger of despotism,
if there were two, they would cut each other's throats,
but there are thirty, and they live in peace and happiness.[11]*

--

*The circumstances of human society are too complicated to be submitted to the rigour of mathematical calculation.[12]*

Diversity is a concept that has to do with differences, and with subgroups. In this chapter, we will discuss a metric for diversity. Although the diversity concept is fundamental in many disciplines, e.g. sociobiology, ecology, social science, biological taxonomy, statistics, and genetics, there are still many unresolved issues. One typical problem is that of selecting which clustering algorithm to use. Another, that the difference metric to use is often not very obvious. There is always a risk that the clustering is made on partly irrelevant factors, and that several important differences are missed.

## 4.1  BEHAVIORAL DIFFERENCE

If we assume that two agents have received the same information in the same kind of situation, they can still react differently. Sometimes, differences are due to obvious technical differences such as mechanical properties of various robot individuals, but most interesting are behavioral differences that are due to different situation assessments, different utility and probability reasoning, different adherence to norms and agreements, different skills, and the impact of different previous experience. How can we objectively compare the behaviors of agents?

### 4.1.1  GENERAL DEFINITION ATTEMPT

The complexity of behavioral difference is not easily captured in a metric. Balch (1998) has suggested the following approach[13]: Take $j$ robots. One such robot can at each time be in one of $i$ different discrete perceptual states. Now, let :

$r_j$ : the robot nr. j
$s_i$ : the perceptual state i
$a^i_j$ : the action selected by robot $r_j$ when it is in perceptual state $s_i$
$\pi_j$ : $a^i_j = \pi_j(s_i)$ ; this is called j's *policy*.
$n^i_j$ : number of times during experiment that $r_j$ has been in $s_i$
$n_{tot}$ : sample size of experiment

---

[11] Voltaire (1694-1778), French philosopher, author. *Letters on England*, Letter 6, "On the Presbyterians" (1732). From The Columbia Dictionary of Quotations.
[12] Marquis de Custine (1790-1857), French traveler, author. *Empire of the Czar: A Journey Through Eternal Russia*, ch. 29 (1843; rev. 1989). From The Columbia Dictionary of Quotations.
[13] The notation has been slightly modified compared to Balch's, in order to generalize perceptual states and not to confuse the attemptive behavioral difference metric with the concept of diversity.

In this thesis, we regard all policies to be activities of a complex nature, such as "move along the side", or "dribble the ball". Therefore, we can not imagine what the notion of difference between two various policies could be other than a check whether the policies are identical or not. Consequently, we define a response difference binary value between robots **a** and **b** in state **i** as : $|p_a(i) - p_b(i)|$, and this formula may assume values 0 if the policies are identical, or 1 if they are different (i.e. there are no grades of difference in this formula). Then two robots, $r_a$ and $r_b$, can be compared by integrating the differences in responses over all possible perceptual states:

$$\Phi_1(a, b) = \frac{1}{n_{tot}} \int |\pi_a(i) - \pi_b(i)| di$$

( where $1 / n_{tot}$ is a normalizing factor to make $0 \leq \Phi_1 \leq 1$ ).

A problem with $\Phi_1$ is that it weighs all differences across perceptual states equally. But the robot might spend more time in certain states, and states more frequently visited should be emphasized. For this purpose, we will use a post facto computation of $n^i_j / n_{tot.}$ and call this value $p^i_j$.

A definition of behavioral difference could then be:

$$\Phi_2(a, b) = \int \frac{(p^i_a + p^i_b)}{2} |\pi_a(i) - \pi_b(i)| di .$$

With a measurement of behavioral difference at hand, Balch makes four definitions, that we will here rename to reduce some ambiguity. Furthermore, we will immediately condense his four suggested definitions into three, and also introduce mathematical notation for the definitions:

**Def.1**
Robots $r_a$ and $r_b$ are *behaviorally equivalent*
$r_a \equiv r_b \Leftrightarrow \forall i: \pi^a_i = \pi^b_i$.

**Def.2**
Robots $r_a$ and $r_b$ are *behaviorally ε-similar*
$r_a \equiv_\varepsilon r_b \Leftrightarrow \Phi_2(r_a, r_b) < \varepsilon$.

**Def.3**
A robot society R is *behaviorally ε-homogenous*
$R^\varepsilon \Leftrightarrow \forall r_a, r_b \in R : r_a \equiv_\varepsilon r_b$

4.1.2 PROBLEMS WITH THE GENERAL DEFINITION ATTEMPT

Balch immediately indicates four limitations with the attemptive definition of behavioral differences:

1. There are some preparations that have to be done with the data sampled. Analog input that the robots receive must be transformed into discrete perceptual states. Complex actions like "walk along wall" should be treated as a nominal value, with response difference set to 0 when the robots use the same policy, and to 1 otherwise.
2. In a realistic experiment there are so many possible perceptual states that the computation above will be very heavy.
3. This approach implicitly assumes that robots use a fixed policy per perceptual state. This is the case for some agent architectures, but certainly not for all. If the robot uses a state machine for situation assessment, the same perceptual state may lead to different actions, depending on the previous events.
4. Two robots might have the same policy for a certain perceptual state, but one might have a slower computer, and perhaps miss a state change (or some other aspect of the assessment). Identical policy in a table does not guarantee identical performance in real life.

We agree that the suggested definition for behavioral difference is impractical because it operates on a very low level (perceptual states), thereby making it a difficult scientific problem to sample and calculate, even for a simplistic robot system. It is not realistic to simplify a complex set of actions into a binary policy value; by doing this abstraction, almost all individual differences are filtered away, and the system might seem more homogenous than it actually is. Furthermore, the suggested definition will, as hinted in point four, miss out several of the most important heterogeneity drivers, e.g. technical differences in software or hardware, individual differences in skills, and activity variations that are due to roles or to coalition membership.

### 4.1.3 AGENTS UNDER SITUATED AUTOMATA CONTROL

Consider agents that use internal situated automata for behavioral control. Now, if one of those agents is in a particular state whenever a snapshot is taken, and that each state also includes performing a unique activity (staying idle is regarding as a special form of activity), then we suggest that an analysis of the state distribution for a number of agents using the same automaton design over some sufficiently long period of time may indicate behavioral differences between the agents, arising from variation of roles and different accumulation of experiences. In the case that the state apart from - or instead of - a normal action can also invoke a special action, and that action is selected, then this is logged for further analysis as if the agent had instead virtually been in another state, symbolizing performance of that special action.

This method does not inspect what perception the agents has had, or assume direct couplings between perception and action, like the policy-based general definition attempt does. Yet, we believe that an insight into behavioral differences can sometimes be reached using this analysis[14]. Obviously, chance may prevent or exaggerate the opportunities for a certain agent to display a certain behavior, but if the state distribution is measured during a long time, the randomness in the environment is minor to the primary or secondary effects of actually different behaviors. See page 30 for more information on situated automata.

Finally, a major part of the reason why one agent may display a certain behavior more often than another agent, or operate with some variation of the behavioral details, often lie in different role assignments, as suggested on page 6. This is a major problem for our suggested behavioral difference measurement, since the effects of role assignment is something that we want to capture in a behavioral difference metric. Imagine two physically identical robots having separate home positions defined, and using different settings for a stamina handling parameter. The state 'move to home position' will then induce similar behaviors, but with variation of the repositioning goal and of the resource management while moving. A behavioral difference metric should take into account the similarities of their activities, as well as the fact that they are moving to different positions, and with different speed. This makes the analysis extremely complex, whereas the intention of movements often is not visible to an outside observer, and primarily because we cannot say how to weigh the similarities against the differences in the observations when summarizing for a behavioral differences metric.

---

[14] For an overview on the Markovian properties of situated automata and Multi-agent Markov Decision Processes, see footnote on page 31.

## 4.2 SOCIAL ENTROPY

Diversity is conceptually coupled not only to behavioral differences, but also to the proportions of subsets that make up the agent society. These subsets are clusterings of agents, and the way they are clustered depends directly or indirectly on behavioral or other important differences.

### 4.2.1 ENTROPY

Entropy has to do with disorder or randomness in a system. An early research field to focus on entropy was thermodynamics. The central equation of statistical thermodynamics, formulated by Ludwig Boltzmann, relates entropy to atomic disorder. If S is the entropy of the thermodynamic system and W is the system disorder, then $S = k \log W$, where k is Boltzmann's constant, $1.38 * 10^{-23}$ J/K, one of the most fundamental physical constants (Halliday and Resnick, 1988).

Later, C. E. Shannon and W. Weaver faced another problem when describing communication systems. They were working at Bell Telephone Laboratories during World War II and were trying to find ways to utilize communication channels to the maximum extent possible (Fiske, 1992), (Shannon and Weaver, 1949). The obstacle is noise, and the remedy a proper amount of redundancy to reduce uncertainties and the entropy in the received messages. For this reason, they wanted to quantify the randomness in an information source, to be able to calculate the minimum bandwidth needed for error-free transmission. Their solution was to define *information entropy* H(X) of a symbol system X in coding theory, which can then be used as a lower bound of the number of bits needed to send a multi-symbol message. If X is a random variable that will assume discrete values in the set
$\{x_1, x_2, x_3 \ldots x_N\}$, and $p_i$ is the proportion of $x_i$, then the information entropy of the symbol system, in bits, is:

$$H(X) = - \sum_{i=1}^{N} p_i \log(p_i)$$

### 4.2.2 THE USA TODAY INDEX OF ETHNIC DIVERSITY

The newspaper USA Today uses an index of ethnic diversity suggested by Meyer and McIntosh (USA Today, 11 April, 1991). It is a probability-based index, ranging from 0 to 1, indicating the probability that two people chosen at random with replacement will *differ along at least one dimension*. A value of zero would therefore apply to a population where everyone was the same on every social dimension. The newspaper uses this index for articles about ethnicity. At least four dimensions are used: nationality, race, religion, and culture. We feel that the scales used for such complex dimensions might be too naive, historically charged with connotations, or at least intellectually humiliating. The option 'Hispanic' in questionnaires is typically vague. But if we want to investigate diversity issues, the USA Today Index has some nice properties, e.g. a crisp conceptual link from diversity to probability, which many people intuitively like.

En passant, the ethnic diversity of the United States as a whole, drawn from the 1990 census, was 40%, according to the USA Today study mentioned above. The most diverse metropolitan area was Los Angeles with 71% (up from a 1980 index of 62%), the least diverse metropolitan area was Dubuque, Iowa, with an index of only 2%.

A drawback with the USA Today index is that it is not recursive. If a society is combined from disjoint sub-societies, then the diversity of the new society should preferably be the weighted sum of the diversities of the sub-societies (the recursive criteria), and this does not hold for the USA Today index.

### 4.2.3 SIMPLE SOCIAL ENTROPY

Many researchers have found that the Shannon & Weaver entropy metric presented above is suitable also in other fields, e.g. in sociobiology, ecology, social science, biological classification, statistics, and evolutionary genetics (Balch, 1998). We will show how easy it is to calculate the simple social entropy for a society if the classes are distinct.

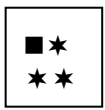

The society Y pictured above is composed of one block and three stars. If $p_1$ is the probability that an element chosen at random will be the block, then we realize that $p_1$, for the block class, has to be 1 out of 4, in other words $p_1=0.25$. In the same fashion, the star class will have $p_2=0.75$.

The entropy for this society is then $H(Y) = -\sum_{i=1}^{N} p_i \log(p_i) = -((p_1 \log_2(p_1)) + (p_2 \log_2(p_2)))$

$= -(0.75 \log_2(0.75) + 0.25 \log_2(0.25)) = 0.811$

As a second example, the entropy for a society built from sub-societies can be derived recursively, as in the following:

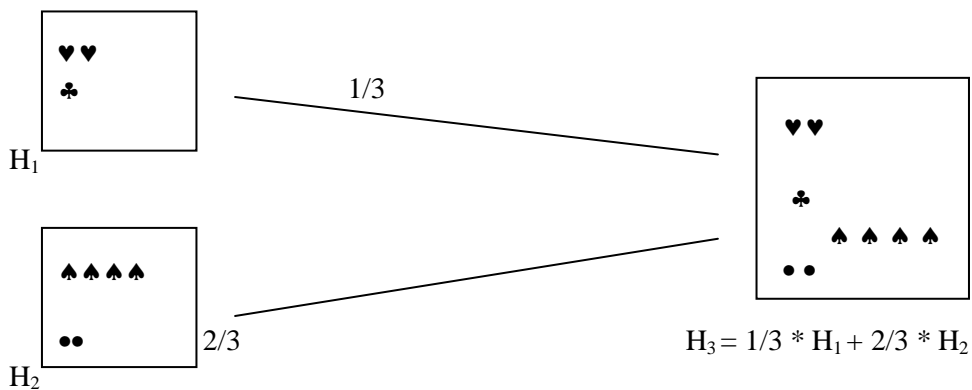

The simple social entropy metric lacks in sensitivity to the degree of difference between the subsets, and a single number does not tell how many classes of agents there are. Many radically different situations therefore end up with the same simple social entropy value.

### 4.2.4 TAXONOMIC DISTANCE

To calculate the taxonomic distance between two elements, we must first put them in a classification space. The classification space is multi-dimensional, with one dimension per trait[15] (e.g. length or color), and the location of each element in this classification space depends on the values for each trait. The taxonomic distance between two elements is the Euclidean distance in the classification space.

### 4.2.5 HIERARCHIC SOCIAL ENTROPY

It is not likely that the agents can be immediately divided into a few distinct subgroups in a binary manner like the simple social entropy metric assumes. The challenge of finding the spatial structure of the agent properties is the same faced by biologists when trying to classify species. Numerical taxonomy is a field of biology concerned with ordering various organisms hierarchically. The approach most widely used, is to draw a taxonomic tree, where strong similarities between organisms are represented by grouping near the leaves of the tree, and weaker similarities are connected by branches closer to the root. In this manner, 1.4 million living species of all kinds of organisms on our planet have been classified (Wilson, 1988)[16].

Before the taxonomic tree can be drawn, the organisms must be clustered within the classification space. We will use a clustering algorithm that will work in the following way:

Use a variable h for taxonomic distance so that when h=0 each organism is a separate cluster, and when h=1 all organisms are grouped together. In between, the number of clusters will decrease as we move from h=0 to h=1. One algorithm with non-overlapping clusters that works this way is the AutoClass clustering program. Note that h is a parameter of the clustering algorithm, so we must vary this parameter in order to find where the clusters integrate with each other. As we increase h from 0 towards 1, the clusters become fewer, thus the simple social entropy decreases. Finally, the entropy reaches zero. In a less diverse society, the drop towards zero entropy is faster than in a more diverse society, since the clustering can more quickly engage other members.

We see that the social entropy value then is a function of the society under evaluation and of the parameter h used for clustering, hence H(R, h) = H(R) at taxonomic level h.

Finally, the hierarchic social entropy is defined as:

$$S(R) = \int_0^\infty H(R, h) \, dh$$

Hierarchic entropy preserve the basic properties of simple social entropy, but also shows scale invariance and accounts for continuous differences between members in the society (Balch, 1998).

## 4.3 BEHAVIORAL DIVERSITY

The ideal would be to have a working metric of behavioral differences as suggested on page 15, then cluster the agents based on their behavioral differences in order to calculate a hierarchic social entropy - this being the behavioral diversity metric. Unfortunately, this way of grouping the agents may be impossible to achieve in practice, for reasons indicated in the previous sections. Therefore, a working general metric of behavioral diversity is yet to be found.

---

[15] Recently, gene data sources are becoming available. Species clustering and evolutionary trees can probably be made based on gene code sequences. (The mutation phenomenon poses a special challenge).

[16] According to Wilson, biologists agree that this identified, named, and catalogued part is only a fraction of all species alive, and that 10 million species is reasonable estimation.

**4.4 CASE-BASED DIVERSITY**

Since we do not expect to be able to measure the behavioral diversity, we will in many cases resign to let a few agent property values, which we have identified as important in the specific case, and that we want to base our study on, to form the taxonomic dimensions. If we have a team of agents where agent property A can vary, and we assume that the way property A is set within a team has an impact on team performance, then we will simply assign property A to be a taxonomic dimension.

# 5 TRANSIENT DIVERSITY

*Ut tensio sic vis* [17]

The diversity of the system can, and most often will, change during execution. There might be a fault in one member of a sub-team, leading it to die and/or to be removed from the scene. The diversity of the team is then immediately affected, since the number of members and the structure of the team changed. Imagine further that the multi-agent system itself could change the settings for many software parameters to achieve some performance effect at any given moment. In a robot system, the individual robot may also change its physical configuration or allow itself to be re-equipped to participate in another robot group.

We therefore realize that a diversity index (calculated by using any of the methods in the previous chapter) is bound to the time where the snapshot of the team was taken. One second later, an agent might have died, and another agent starting to change its group membership. With no new domain events or changes in the environment, e.g. no adversarial teams, no faults, etc., the multi-agent system could theoretically continuously perform with a stable, or harmoniously varying diversity index. However, we will particularly study systems where there is a risk of fatal component faults and a display of adaptive behavior, thereby varying the diversity index in interesting patterns over time.

There is a need for a concept that captures the variations in the diversity over time, the flexibility for increased or decreased diversity that is inherent in the system, and how this flexibility may be utilized in a rational way to increase teamwork efficiency according to situations encountered. Our concept of transient diversity is intended as a move in this direction.

---

[17] Robert Hooke (1635-1703) presenting his law of mechanical vibrations in 1678. Roughly: 'as the force so is the displacement'.

## 5.1 DIVERSITY CHANGE AND STABILIZATION

The diversity of the multi-agent system can suddenly change, or gradually adapt, due to external events or internal decisions. In fact, most the diversity drivers presented earlier in this thesis are dynamic throughout execution. In general, this change is distinct, and after the change, the system stabilizes at the new diversity level. Consider the case when an agent malfunctions, and the system adapts to this loss. If the malfunctioning agent had a special role or special skills, we quickly realize that the diversity of the system probably dropped when that special agent became inoperative. The adaptive behavior of the system might partly compensate for the loss, but in this example probably not reach the exact same diversity level as before the malfunction. E.g., another robot might physically reconfigure to undertake the functions of the malfunctioning robot. The graph below illustrates this example.

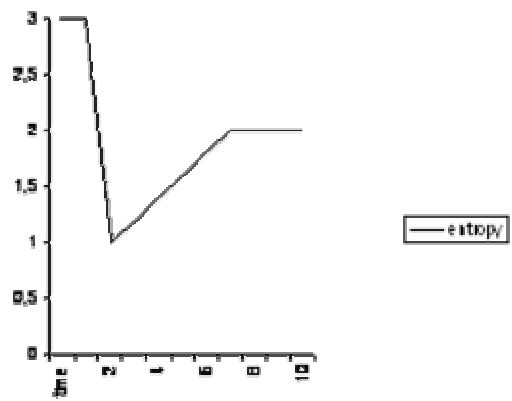

*Figure 1: One robot malfunctions, but the remaining team partly compensates for the loss.*

## 5.2 DIVERSITY VIBRATION

*Society never advances.*
*It recedes as fast on one side as it gains on the other...*
*Society acquires new arts, and loses old instincts.*[18]

Imagine a multi-agent system with inherent capacity of adapting its diversity according to environmental or internal factors. Now, as the system adapts to a change, there might be several primary or secondary effects, some wanted, and some undesired. These effects will maybe induce the system to begin a new adaptive action that more or less reset the original one, or even compensate back beyond the original mode. Sometimes, the action, the direct and secondary effects, and the reaction, might theoretically put the system in an oscillating mode. When does this happen, and what mathematical model governs such an oscillating system behavior?

We suggest that *diversity vibration* is a phenomenon governed by differential equations. Furthermore, in some rare cases, there might be periodic environmental factors that have a frequency or a force amplitude that harmonize in a special way with the adaptive capacity of the multi-agent system. These factors will then quickly induce an over-reactive oscillating behavior, greatly reducing system performance.

Many significant problems in engineering, physics, and the social sciences have been described using derivatives of the unknown function. Such equations are called differential equations. This section will discuss a possible framework for a virtual diversifying force which we, in a modeling sense, consider imposed on a multi-agent system, causing the system to diversify or homogenize.

We will use the name *diversification* for the concept of increasing or decreasing diversity in the system, and the corresponding metric for diversification will be signed accordingly. If D(t) is the diversity of the system at time t, then we define the *diversification speed* to be D´= dD/dt.

### 5.2.1 DIVERSIFYING FORCE INTRODUCED

Consider the case of the team of agents effected by a number of diversity drivers, as presented earlier in this thesis. Now, assume that the combined effect of all those drivers taken together, is a *diversifying force*. If there were no heterogeneity drivers (of if the homogeneity drivers would efficiently cancel the overall societal effects of the heterogeneity drivers), there would be no diversifying force, and the system would have a diversity of zero. We will define the diversifying force as a summation of three components; overcoming 'inertia', 'resilience', and 'resistance'.

### 5.2.2 INERTIA

At the onset of a diversifying force, we would like to reflect the fact that there often is inertia before the full speed of diversifying can be reached. Let us regard the team of agents as an object with a property called mass. The mass property will be defined as a linear coupling between diversifying force and the rate of change in diversification speed. We want to reach a formula where a rapid acceleration (or deceleration) of diversity in either direction will be reflected in a strong diversifying force. Similarly, a high mass would need a strong diversifying force to obtain a specific acceleration of diversity. Using M for the mass of the system, if the diversity is D, the rate of change in diversifying speed is D´´, then we define the needed diversifying force as:

$$F_C(t) = M * D´´.$$

---

[18] Ralph Waldo Emerson (1803-82), U.S. essayist, poet, philosopher. Essays, *Self-Reliance* (First Series, 1841).

### 5.2.3 RESILIENCE

Imagine that there is a sudden event that changes the diversity of the system, but that there are immediate compensatory actions in one or more of the diversification drivers, and that the system itself or the surrounding environment therefore will, gradually, try to move the system back towards the heterogeneity that was present before the event. In a language borrowed from physics, we can say that the total diversifying force changed after the event because of an *resilience* in the system, in order to bring the system back towards its previous equilibrium.

If D is the diversity, the resilience of the system is E, and we assume a linear relationship, then we will express the restoring component of the diversifying force as :

$$F_E(t) = E * D.$$

### 5.2.4 RESISTANCE

There is a limit to how much a multi-agent system can normally change and adapt in a limited amount of time. It is easy to realize that in all realistic systems there are <u>damping</u> effects that counteract quick series of rapid changes in the system.

This damping can be due to technical bottlenecks such as bandwidth, available processing resources, or latency in communication throughput for coordination. In robot systems, it may take a certain amount of time for the robots to reconfigure their physical appearance, or activate various mechanical modules. The damping property can also partly be specified in the agent software, to avoid too quick adaptive actions.

Notice that the diversity metric concerns the functioning parts of the system, so maliciously killing half of the agents in a split second by override intervention will presumably have effects on the diversity of the remaining team so drastic that the team itself would in normal operation not be able to reproduce such a change in such a short time.

Assume that a change back toward homogeneity is damped in the same amount as a previous change towards heterogeneity was. We will define the system *resistance* R, and the associated diversifying force component overcoming that amount of resistance for a specific diversification speed D´ as:

$$F_R(t) = R * D´.$$

### 5.2.5 DIVERSIFYING FORCE DEFINED

We will define the diversifying force at each time t to be the addition of the mentioned three components. Hence:

$$F(t) = F_C(t) + F_R(t) + F_E(t) = M*D'' + R*D' + E*D$$

### 5.2.6 UNDAMPED FREE DIVERSITY VIBRATION

If there is no damping and no diversifying force acting on the system, then the equation for diversifying for presented in the previous section reduces to:

$$MD'' + ED = 0,$$

which has the solution[19]

$$D = A\cos(\omega_0 t) + B\sin(\omega_0 t),$$

where $\omega_0$ is the natural circular frequency given by $\omega_0^2 = E/M$. The constants A and B can be determined via the initial conditions. One usually rephrases this solution into:

$$D = r \cos(\omega_0 t - \delta),$$

by using a phase angle $\delta$ given by $\tan\delta = B/A$ and an amplitude r as $(A^2+B^2)^{1/2}$. Because of the periodic property of cosine, we see that the natural period for the simple harmonic motion is $T = 2\pi/\omega_0$. The amplitude represents the maximum displacement of the diversity from its equilibrium.

### 5.2.7 DAMPED FREE DIVERSITY VIBRATION

The differential equation describing a damped free diversity vibration is:

$$MD'' + RD' + ED = 0.$$

The roots of the characteristic equation will give us three distinct cases. In all cases, the oscillation will die out over time, due to the resistance. The first case is called *overdamped*, the second *critically damped*, and the third *underdamped motion, or damped vibration*.

In the overdamped and critically damped cases, the diversification will creep into an equilibrium. Depending on the initial conditions, the motion may at first overshoot the equilibrium. In the damped vibration, we will see something that resembles a cosine curve with decreasing amplitude. This vibration is not truly periodic, but there is a quasicircular frequency, and a quasiperiod.

### 5.2.8 FORCED UNDAMPED DIVERSITY VIBRATION

If there is a periodic diversifying force active, say $F_0\cos(\omega_0 t)$, then the multi-agent system will display a diversity vibration given by the equation

---

[19] For the details of solving differential equations, see e.g. (Boyce and DiPrima, 1986).

$$MD'' + ED = F_0\cos(\omega_0 t).$$

The vibration is a sum of two periodic motions of different frequencies and amplitudes. Trigonometric analysis shows that we have two interesting situations. In the first situation, the two motions have the same amplitude, and amplitude modulation between the motions results in a slowly varying sinusoidal amplitude, called a *beat*.

In the second situation, the period of the forcing function is the same as the natural period of the system, then the motion becomes unbounded as t increases. This special phenomenon is well known, and is called *resonance*. Diversity resonance can create serious difficulties in a multi-agent system, the instabilities can lead to a rapid disintegration of teamwork and possibly total malfunction of the team.

5.2.9 FORCED DAMPED DIVERSITY VIBRATION

If we add damping to the forced vibration discussed above, the vibration will be given by:

$$MD'' + RD' + ED = F_0\cos(\omega_0 t).$$

After a rather lengthy computation[20], the solution to the forced damped diversity vibration can be found:

$$D = D_t(t) + D_s(t)$$
$$\text{where}$$
$$D_t(t) = R_1 e^{gt} + R_2 e^{ht}$$
$$D_s(t) = F_0 * \cos(\omega_0 t - \delta) / (M^2(\omega_0^2 - \omega^2) + c^2\omega^2)^{1/2}$$

The constants g and h used in $D_t(t)$, are the roots of the characteristic equation. Both $e^{gt}$ and $e^{ht}$ approach zero when t goes towards infinity, hence $D(t) \to D_s(t)$ as $t \to \infty$.
For this reason, $D_s(t)$ is often called the *steady state solution*, and $D_t(t)$ is called the *transient solution*.

5.2.10 TRANSIENT AND STEADY-STATE RESPONSE TO A SINUSOIDAL FORCE

In the plot below, the forcing frequency is selected so that the system is being driven below resonance. The harmonic curve, starting with F(0)=1, shows the diversifying force, and the other curve (the one starting with D(0)=0) shows the diversity displacement. After the transient motion decays and the oscillation settles into steady state motion, the diversity displacement is in phase with force. Notice that the frequency of the steady state motion of the mass is the driving (forcing) frequency, not the natural frequency of the system.

---

[20] For the details of solving differential equations, see e.g. (Boyce and DiPrima, 1986).

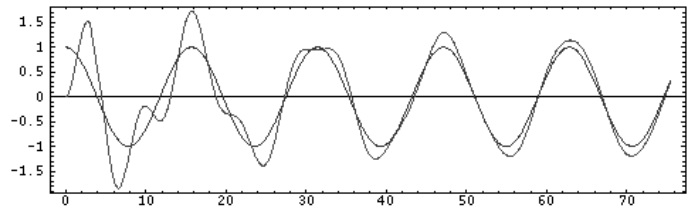

*Figure 2: Transient and steady-state response*

5.2.11 RESONANCE

In the plot below, the forcing frequency is selected so that the system is being driven very near resonance. The curve with the same low amplitude shows the diversifying force, and the curve with increasing amplitude shows the diversity displacement in response to the diversifying force. Since the system is being driven near resonance the amplitude quickly grows to a maximum. Note that after the transient motion decays and the system settles into steady state motion, the displacement lags 90° out of phase with force.

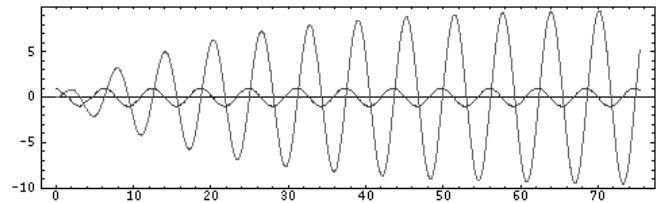

*Figure 3: Resonance*

## 5.3 ANALOGIES TO OTHER SYSTEMS

The theory of linear second order differential equations with constant coefficients have several applications. We have been inspired by the use of differential equations in mechanical systems and electrical networks. In the following table is an overview of the analogy between mechanical, electrical, and diversity vibrations:

| Mechanical System | Electric Circuit | Diversity in a Multi-Agent System |
|---|---|---|
| Displacement, u | Charge, Q | Diversity, D |
| Velocity, u´ | Current, I=Q´ | Diversification speed, D´ |
| Mass, m | Inductance, L | Mass, M |
| Damping, c | Resistance, R | Resistance, R |
| Spring constant, k | Elastance, 1/C | Resilience, R |
| Impressed force, F(t) | Impressed voltage, E(t) | Diversifying force, F(t) |
| mu´´+ cu´+ ku = F(t) | LQ´´+RQ´+(1/C)Q=E(t) | MD´´+RD´+ED=F(t) |

**5.4 TRANSIENCE PREVAILS**

The transient solution to the forced damped diversity vibration allows us to impose initial conditions on a differential equation governing the diversity of a system. The damping will dissipate the diversifying force properly, and the vibration will eventually reflect the sustained response of the system to the diversifying force. However, the multi-agent system will probably seldom have time to stabilize in a steady state diversity vibration, since the evanescent environment will quickly impose new external and internal diversifying forces. This is of course also the case also for other more direct changes in diversity that are not classified as forced damped vibration. Immediately after a change, there are a number of codependent factors that arise, effecting the diversity. Therefore, as we see it plainly appear, the diversity of an agile multi-agent system situated in a dynamic environment, will mostly be in a transient phase. Transience is the quality of being transient, thus, transience will often prevail.

# 6 EXPERIMENT

## 6.1 EXPERIMENTAL PLATFORM

This section will provide an overview of the UBU RoboCup Team, which will be used as an experimental platform (Kummeneje et al, 1999), (Boman, Kummeneje, Lybäck, and Younes, 1999). After an introductory overview, the top-level situated automaton will be described. Although important, the reasoning under uncertainty and the decision theoretic parts of UBU are omitted from this presentation, since decision theory is not a central subject in this thesis.

### 6.1.1 ARCHITECTURE OVERVIEW

The UBU RoboCup Team consists of eleven autonomous processes that represent the goalkeeper and the ten field players. In the future, we might also implement a special twelfth process, called coach. The program system is object-oriented and written entirely in Java, using concepts like multi-threading, encapsulation, polymorphism, listener, and events. Relying on threads in the Java language, the system is also prepared to fully utilize multi-processor computers. Each player agent includes several concurrent threads.

### 6.1.2 MAIN DESIGN IDEAS

The design consists of two main levels. The lower levels are focused on perceiving the environment, and affecting it. The higher levels try to classify the situation and make decisions on what to do. The team uses a semi-complex situated automaton to create a mix between reactive and deliberative design principles. The whole architecture is strongly inspired by the OSI-model (Tanenbaum, 1996). By using a layered approach, the low-level details are isolated from high-level processes. Thus, it will be possible to e.g. exchange the pre-defined soccer server layer, by making a revision to the primitive action layer only.

### 6.1.3 THE SITUATED AUTOMATON

In UBU, the situated automaton is the very tip of the iceberg. Situated automata are state machines linked to the environment via causal dependencies, providing a reactive control mechanism (Russell and Norvig, 1995). Due to the links, the situated automata implicitly carry information about the world. The internal states reflect the presence of some environmental conditions, or, in our case, how those conditions are interpreted. In a particular set of circumstances, the situated automaton is restricted to one behavior, which means that the automaton in one way makes the system inflexible, since the change of internal state should be due only to the change in the external world.

The design of a state machine is fixed throughout program execution, and the states are specific to a certain domain. Consequently, for these reasons, a situated automaton can only work in the environment it has been designed for: it can simply not be expected to operate in a wide variety of unknown circumstances, and thereby it scales poorly for other challenges. The great advantage of a state machine is its excellent real-time performance: static facts are compiled into the structure of the machine, which saves valuable execution time.

The UBU situated automaton has a special design feature: in each state there are two types of possible outcomes: reactive responses and deliberative decisions. Some of the reactive behaviors are mandatory in the RoboCup environment, e.g. correctly interpreting and obeying the messages from the referee. Therefore, the reactive responses to such messages are implemented to supersede deliberation. This is made in the following way: the reactive parts of the current state are evaluated first. If one of the reactive outcomes includes a mandatory state transition, this transition is performed and the automaton goes for its next cycle. If there are no mandatory reactive transitions to perform, and circumstances permit, the deliberative parts of the current state are evaluated, and the activity with the highest expected utility is selected and performed. As with reactive activities, the deliberative activities can also be combined with a state transition, where a special transition is to remain in the same state.

Note the difference between activities and actions in our architecture: we use the term action for the primitive actions, while activity is reserved for a more complex set of conditions and actions. Note also the difference between the situated automata and the subsumption architecture (Brooks, 1989), both known to provide real-time performance. In the subsumption architecture, the decisions are fully Markovian[21], but since the situated automata use internal states, the decisions in UBU are not totally ahistoric.

According to Mataric (1995), using a learning algorithm based on Markovian models, such as a state automaton, will in general unfortunately fail for robots, since the world does not necessarily change neatly in response to agent actions. This is said to be due to the agent's incomplete knowledge of the world, the limited perception, and the uncertainties involved. Hence, the world and the agent cannot both be conceptually integrated into a simple synchronized finite state automaton.

On the other hand, Boutilier (1999) argues that decision and coordination problems can be modeled as a Markov Decision Problem[22], or rather the extension called Multiagent Markov Decision Process[23], and that although agents can not predict which state they will reach when an action is taken, they can still detect the state when it is actually reached.

We feel that Boutilier's argument holds, but that Mataric is partly right in that the uncertainties complicates learning.

---

[21] Place does not permit a full description of Markovian properties, but the concept can partly be described as follows:
A general random Markov process is defined as:

$$P[\, X_{t(n+1)} = x_{n+1} \mid X_{t(n)} = x_n, X_{t(n-1)} = x_{n-1}, \ldots, X_{t1} = x_1 \,] = P[\, X_{t(n+1)} = x_{n+1} \mid X_{t(n)} = x_n \,].$$

In a general discrete time Markov chain, the process makes the transition from the current state to another state with a geometric distribution of time spent in state. Moreover, in a general Semi-Markov process, the time between transitions obey an arbitrary probability distribution (Reichert and Maguire, 1994).

[22] A fully observable Markov Decision Process, **MDP**=<**S**, **A**, **P**, **R**>, comprises the following components:
First, a finite set of states **S** of the system, and a finite set of actions **A** with which the agent can influence the system state. Furthermore, let the probability that action a when executed at state $s_i$ induces a transition to state $s_j$, be called $P(s_i, a, s_j)$, and the system dynamics then given by $\mathbf{P} : \mathbf{S} \times \mathbf{A} \times \mathbf{S} \rightarrow [0, 1]$.
Finally, let $\mathbf{R} : \mathbf{S} \rightarrow \mathbf{R}$ be a real-value, bounded reward function.
It can be shown that an optimal policy can constructed via value iteration over a t-state-to-go value function $V^t$ by setting

$$V^0(s_i) = R(s_i) \text{ and } V^t(s_i) = R(s_i) + \max_{a \in A} \left\{ \beta \sum_{s_j \in S} P(s_i, a, s_j) V^{t-1}(s_j) \right\}.$$

Use $\beta=1$ (no discounting) for finite horizon problems. For infinite horizon problems, the value iterations converges to the optimal value function $V^*$. See Boutilier (1999) for more details.

[23] Actions are distributed among multiple agents: let $\alpha$ be the finite collection of n agents with each agent having at its disposal a finite set $A_i$ of actions. Then define the *Multi-agent* Markov Decision Process **MMDP**= <$\alpha$, **S**, $\{A_i\}_{i \in \alpha}$, **P**, **R**>.
It can be shown that the *optimal joint value function* $V^\sim$ can be computed via joint policies if the agents do not have competing interests. The MMDP can be regarded as a type of stochastic game (Boutilier, 1999).

## 6.2 RESULTS

*We shall not cease from exploration*
*And the end of all our exploring*
*Will be to arrive where we started*
*And know the place for the first time.* [24]

The performance of a complex system, such as the Ubu RoboCup Team, is a result of the interaction dynamics, both within the system, and with the environment where the system is situated.

The metric chosen in this thesis for evaluating the performance is goals scored in games against a control team, subtracted by the number of scores that the control team made. Although scores may be a simplistic measurement, it summarizes whether the agents meet their overall ambition of winning the game. We have made experiments on variation of role assignment and trait settings, typical heterogeneity factors, and measured the impact on team performance.

### 6.2.1 EXPERIMENT ON ROLE AND TRAIT VARIATION

The following table shows how roles have been set differently in 5 different teams, and in the control team.

| Team | # Goalie | # L Defender | # C Defender | # R Defender | # L Midfield | # CL /C Midfield | # CR Midfield | # R Midfield | # L Forward | # C Forward | # R Forward |
|---|---|---|---|---|---|---|---|---|---|---|---|
| Kids0 | 1 | | | | | 10 | | | | | |
| Agr | 1 | | 3 | | | | | | | 8 | |
| Kids2 | 1 | | 5 | | | | | | | 5 | |
| Kids1 | 1 | 1 | 1 | 1 | 1 | 1 | 1 | 1 | 1 | 1 | 1 |
| Kids3 | 1 | 1 | 1 | 1 | 1 | 1 | 1 | 1 | 1 | 1 | 1 |
| Control | 1 | 1 | 1 | 1 | 1 | 1 | 1 | 1 | 1 | 1 | 1 |

*Table 1: Player Roles*

Each team competed against the control team in three consecutive games, and the score difference is noted as a performance metric. From the table above, we can readily see that Kids0 have only two possible home positions that the players use during the game, one for the goalie and one for the other players (home position is an important part of the role of the soccer agent). We also see that Agr and Kids2 have two, and the rest have eleven home positions.

Using the simple social entropy presented on page 19, we can calculate the entropy of the team with respect to the positioning property. This entropy ranges from 0.44 for Kids0, up to 3.46 for the last three teams analyzed.

---

[24] T. S. Eliot (1888-1965), Anglo-American poet, critic. Little Gidding, pt. 5, in Four Quartets. From The Columbia Dictionary of Quotations.

Another setting that is being varied is the trait of offensiveness. In Kids0 and Kids3, all the players use a setting of 80%, in Kids1 50%, and in Agr 99%. The Kids2 team use 50% for defenders and 90% for forwards, and the control team uses 30% for defenders, 60% for midfielders, and 90% for forwards.

The findings are summarized in the table and diagrams below.

| Team name | Number of player positions (incl. Goalkeeper) | Average team setting for offensiveness trait | Cluster offensiveness settings | Offensiveness clusters for field players |
|---|---|---|---|---|
| Kids0 | 2 | 80 % | 0.8 | 1 |
| Agr | 3 | 99 % | 0.99 | 1 |
| Kids2 | 3 | 97 % | 0.5, 0.9 | 2 |
| Kids3 | 11 | 80 % | 0.8 | 1 |
| Control | 11 | 55 % | 0.0, 0.3, 0.6, 0.9 | 3 |
| Kids1 | 11 | 50 % | 0.5 | 1 |

*Table 2: Settings*

| Team name | Entropy positioning | Entropy offensiveness trait |
|---|---|---|
| Kids0 | 0,439497 | 0 |
| Agr | 1,18872 | 0 |
| Kids2 | 1,3485879 | 1,3485879 |
| Kids3 | 3,459432 | 0 |
| Control | 3,459432 | 1,8676 |
| Kids1 | 3,459432 | 0 |

*Table 3: Calculated Entropies*

| Team name | Performance against Control (three games) |
|---|---|
| Kids0 | -15 |
| Agr | -9 |
| Kids2 | -7 |
| Kids3 | -1 |
| Control | 0 |
| Kids1 | -8 |

*Table 4: Performance Results*

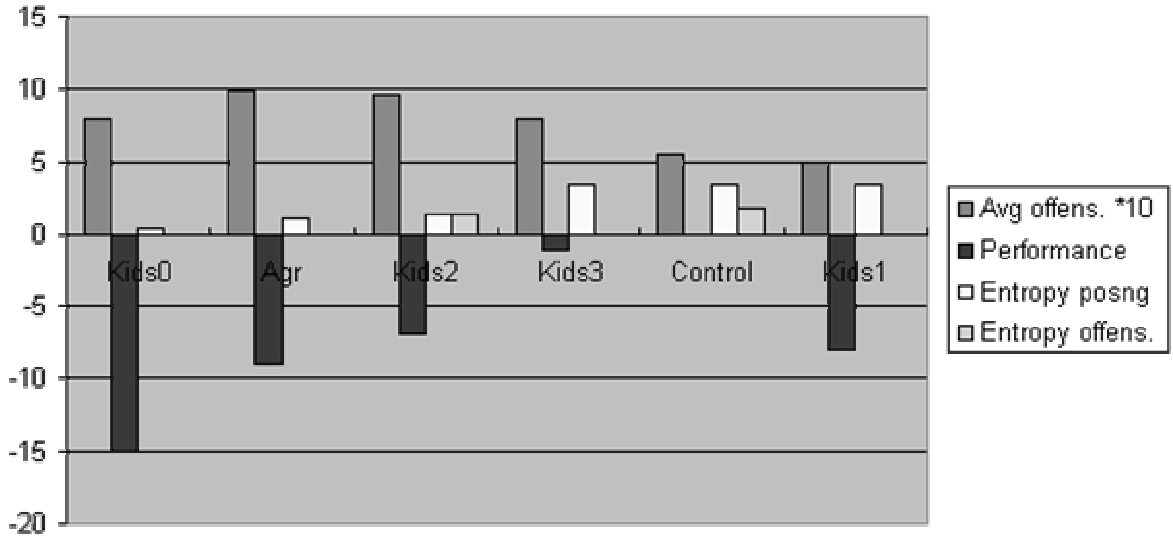

*Figure 4: Diagram of positioning entropy and performance, with additional bars for average offensiveness and its entropy*

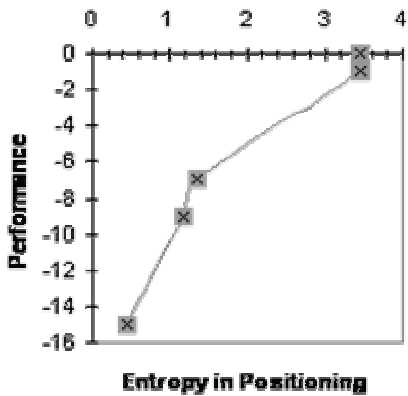

*Figure 5: Diagram of positioning entropy versus performance. When the maximum entropy is reached (all players have unique positions), performance can still be improved by adjusting the offensiveness settings. One of the teams, Kids1, was obviously not well designed: on the one hand it has proper positioning yielding a high entropy, but the performance was quite mediocre due to a too low offensiveness setting. For this reason, Kids1 is not included in this graph.*

## 6.3 DISCUSSION

*Science may be described as the art of systematic over-simplification.*[25]

We have focused on the simulator league of RoboCup as an experimental domain. However, we expect most of these results to apply *mutatis mutandis* to other settings.

Generally, experiments in this field suffer from the problem of mapping experimental parameters cleanly into relevant theoretical concepts. Consequently, effects attributed to one theoretical concept could (at least partly) actually be due to another concept. According to Cohen, Riolo, and Axelrod (1998), this "confounding of potential theories" is very common in multi-agent systems research, partly because it is difficult to build models of social mechanisms.

We believe that the experiment performed in any case show that there is a positive correlation between heterogeneity in the positioning of *soccer* agents (a role display), and team performance. A proper variation on the offensiveness setting (a trait), also seem to have a positive effect on performance. Time unfortunately does not allow a more comprehensive experiment with more variants, or experiments on the transient diversity concept.

---

[25] Karl Popper (b. 1902), Anglo-Austrian philosopher. Quoted in: Observer (London, 1 Aug. 1982).

# 7 CONCLUSIONS

## 7.1 SUMMARY

This thesis presents a systemization of various factors that clearly influence the diversity of a multi-agent system. It points out some areas of interest, presents an experiment on diversity in a synthetic soccer team, and speculates on various aspects of the diversity concept.

The research area of diversity in multi-agent systems contains several hypotheses, whereof many have been presented in this thesis. Scientific credibility is not always convincingly established, since many theories herein are difficult to define operationally for empirical measurement. A major challenge facing multi-agent theory is how suggested, functionally significant, concepts can be translated into procedures for systematic investigation and verification. Empirical data must therefore be found via various methods, these data must then be thoroughly analyzed, and the associated theories evolve accordingly, or be consigned to oblivion.

We are assured that continued research in diversity management for multi-agent systems can make immediate contributions to existing agent-based applications as well as to long-range developments in future multi-robot technologies.

## 7.2 FUTURE RESEARCH

### 7.2.1 DIVERSITY METRICS AND TEAM PERFORMANCE

As indicated on page 15, the diversity metric available today has clear drawbacks. A practical and theoretically sound way to measure heterogeneity at any given point in time is of utmost importance to any study within agent team diversity research.

It is not always the case that a heterogeneous team will perform better than a homogenous one. The optimal design and setting of several of the diversity drivers is coupled to the task environment. To find working methods to make the team itself understand the task environment and adjust the diversity accordingly, would greatly benefit the team's performance in a broader task environment.

### 7.2.2 SUBJECTIVE DIVERSITY

In analogy with the subjective aspect of coalitions, the decision base for the agents' autonomous adaptive behaviors that adjust diversity driver values may in part be grounded on subjective evaluations of team performance and team diversity. The effects of subjectivity when evaluating performance and diversity, and associated metrics to these phenomena, are open research issues.

### 7.2.3 TRANSIENT DIVERSITY

The suggested concept of transient diversity has yet to be investigated in an implementation. There is a need for empirical validation through measurements of diversity transience.

### 7.2.4 THE RESCUE CHALLENGE

The RoboCup Rescue Challenge, as suggested by (Kitano et al, 1999), will promote research in the domain of large numbers of heterogeneous agents acting in a real-time hostile environment under severe time pressure.

# ACKNOWLEDGEMENTS

First, this project would not have been possible without the academic leadership by Magnus Boman (thesis supervisor). Second, the project would certainly not have been equally enjoyable without the energetic work and always-friendly support by Johan Kummeneje. The project has also benefited from the work and support by Håkan L. Younes, Harko Verhagen, and Christian Guttmann.
Finally, I acknowledge and thank the Scandinavia-Japan Sasakawa Foundation, the Royal Institute of Technology and Stockholm University for supporting my study journey to Japan, and my registration at the IJCAI´99 conference.